\documentclass[10pt,journal,onecolumn]{IEEEtran}
\usepackage{cite}
\usepackage{amsmath}
\usepackage{graphicx}
\usepackage{algorithm}
\usepackage{algpseudocode}
\usepackage{multirow}
\usepackage[table,xcdraw]{xcolor}
\usepackage{authblk}
\usepackage[bookmarks=false]{hyperref}
\usepackage{xcolor}

\algblock{Input}{EndInput}
\algnotext{EndInput}
\algblock{Output}{EndOutput}
\algnotext{EndOutput}
\newcommand{\Desc}[2]{\State \makebox[2.4em][l]{#1}#2}

\newcommand*{\affaddr}[1]{#1} 
\newcommand*{\affmark}[1][*]{\textsuperscript{#1}}
\newcommand*{\email}[1]{\textit{#1}}

\begin{document}

\title{Fake Reviews Detection through Ensemble Learning}

\author{%
Luis Gutierrez-Espinoza\affmark[1], Faranak Abri\affmark[1], Akbar Siami Namin\affmark[1], Keith S. Jones\affmark[2], and David R. W. Sears\affmark[3]\\
\affaddr{\affmark[1]Department of Computer Science,}
\affaddr{\affmark[2]Department of Psychological Sciences,}
\affaddr{\affmark[3]Performing Arts Research Lab}\\
\affaddr{\affmark[]Texas Tech University}\\
\email{\{Luis.Gutierrez-Espinoza, faranak.abri, akbar.namin, keith.s.jones, david.sears\}@ttu.edu}\\
}

\maketitle

\begin{abstract}
    Customers represent their satisfactions of consuming products by sharing their experiences through the utilization of online reviews. Several machine learning-based approaches can automatically detect deceptive and fake reviews. Recently, there have been studies reporting the performance of ensemble learning-based approaches in comparison to conventional machine learning techniques. Motivated by the recent trends in ensemble learning, this paper evaluates the performance of ensemble learning-based approaches to identify bogus online information. The application of a number of ensemble learning-based approaches to a collection of fake restaurant reviews that we developed show that these ensemble learning-based approaches detect deceptive information better than conventional machine learning algorithms.  
\end{abstract}


\section{Introduction}

There are several factors that contribute to the success or failure of a business, and customer satisfaction is always listed as one of the most important ones. Customer satisfactions and good reputation is a matter of life and death for many businesses. It is important to take customer's reviews into account seriously and address their concerns concisely. However, online reviews can also be abused by adversaries for different reasons. 
It is therefore important to distinguish between genuine and dubious reviews. 
The automatic detection of fake online reviews is an inherent instance of a simple binary classification problem. As a result, conventional machine learning-based classification or simple statistical-based approaches can be applicable to this problem. Given the natural language-based nature of reviews and the hidden features that might be latent for explicit modeling, the accuracy might vary. 

The use of the most accurate classification algorithms available for deception is essential in order to design the best possible detection platforms. Conventional machine learning algorithms have been extensively used in the literature concerned with the detection of deceptive and/or fake reviews. These conventional approaches offer relatively accurate results for the underlying classification problem \cite{Crawford2015}. More notably, support vector machines (SVM) are frequently reported as the best classifier for a wide range of application domains. However, recent studies are demonstrating the efficacy of ``ensemble learning'' approaches \cite{EnsembleMachine}, which have been shown to outperform conventional machine learning approaches such as SVM \cite{RFvsSVM}. In particular, recently developed ensemble learning-based approaches such as gradient-boosting machines are demonstrating very promising results for  classification. 

Motivated by the recent research on ensemble learning-based approaches to classification problems \cite{PangXN16, AbriSKSN19}, this paper investigates and compares the performance of these ensemble-based techniques with conventional machine learning algorithms. More specifically, ensemble learning-based techniques (e.g., bagging and boosting) are integrated into conventional Support Vector Machines and extreme gradient-boosting machines and are compared with the conventional algorithms. 

To accomplish this experimental study, our research team created a repository, called hereafter the ``{\it Restaurant Dataset,}" for which a group of undergraduate students created fake reviews for three restaurants. The results of our experimental study, conducted on a repository of fake and deceptive reviews created by our research team, validate the hypothesis that the ensemble learning-based approaches outperform their conventional machine learning counterparts. The key contributions of this research are as follows:

\begin{itemize}
    \item Introducing a new dataset on fake reviews, called the ``{\it Restaurant Dataset}.'' 
    \item Investigating the performance of ensemble learning-based approaches to the problem of identifying fake reviews.
    \item Reporting the results of experiments that compare the performance of ensemble learning-based approaches using different machine learning techniques for classification.    
\end{itemize}

This paper is organized as follows: Section \ref{sec:goals} highlights the goals of this research work and the objectives. Section \ref{sec:relatedwork} reviews the literature. A brief technical background of ensemble learning and the classifiers studied is presented in Section \ref{sec:compared}. The experimental setup of the study is discussed in Section \ref{sec:setup}. The results of the experimental study are presented in Section \ref{sec:results}. Finally, Section \ref{sec:conclusion} offers conclusions and sheds some light on future work.

\section{Research Questions and Objectives} 
\label{sec:goals}

The objectives of this paper are two-fold: 1) introducing a newly produced dataset of fake reviews, called the ``{\it Restaurant Dataset},'' and 2) investigating the performance of ensemble learning-based methodologies in comparison with conventional machine/deep learning approaches to the fake-review detection problem. More specifically, this paper addresses the following questions:

\begin{enumerate}
    \item Are standard forms of machine/deep learning algorithms effective in the context of fake review identification?
    \item Can ensemble learning-based approaches improve the accuracy rate for this problem?
    \item How much would the accuracy be improved through hyperparameter optimization? 
\end{enumerate}

\section{Related work}
\label{sec:relatedwork}

Fuller et al.~\cite{Fuller2011} developed three classification techniques -- artificial neural networks, decision trees and logistic regression -- along with an information fusion-based ensemble method for automated deception detection. They extracted 31 text-based deception features (cues) from the labeled statements of crime scenarios. They reported 74\% accuracy with their fusion based model, 73.46\% with the artificial neural network, and 71.60\% with the decision tree model, respectively. 


Xu et al.~\cite{Xu2015} introduced their approach on deception detection called ``Unified Review Spamming Model'' (URSM). The approach is a unified probabilistic graphical model for deception detection in reviews from three online services (Amazon, Yelp and TripAdvisor). By using the hierarchical Latent Dirichlet Allocation (hLDA) model~\cite{Blei2010}, the model ranks the reviews, the reviewers, and the offers based on the degree of deceptivity. They compared this model with human judgments and a classifier trained using \textit{n}-gram features. They achieved higher performance such as 78.8\% accuracy for a TripAdvisor dataset compared to other methods.

Conroy et al.\cite{Conroy2015} provided a survey for deception detection in the context of implementing fake news detecting systems. They focused on two main methods: the linguistic cue approach and network analysis approach.
Their study showed that the performance of linguistic approaches is topic-oriented and successful in some domains. Also, the performance of network approaches varies from average to high based on the knowledge-base which was used. They argued for an approach that combines linguistic cue and machine learning with network-based behavioral data, and  explained the features for implementing such a system. 

Crawford et al.~\cite{Crawford2015} conducted a study on deception detection in reviews using supervised, unsupervised, and semi-supervised techniques. First, they investigated different approaches for feature extraction and divided them into ``review-centric,'' which uses the content of the review, and ``reviewer-centric,''  which uses review metadata related to the reviewers. 

Levitan et al.~\cite{Mendels2017} compared different machine learning techniques with various feature sets and proposed a hybrid deep neural network approach which uses a combination of audio and textual features (spectral, acoustic-prosodic, and lexical) for deception detection. They utilized a subset of the Columbia X-Cultural Deception (CXD) Corpus to evaluate their models and achieved F1-score of 63.90\% for their deep-hybrid model and precision of 76.11\% for their random forest model. In their next work~\cite{Levitan2018}, they examined automatic deception detection in the interview scenario. They used different sets of features: semantic features collected using LIWC, linguistic featured collected from previous study, personal traits (gender and native language) and also emotional level and detail level scores collected using  Dictionary of Affect Language (DAL). Applying three different classifiers (Random Forest, Logistic Regression, and SVM), they achieved 72.74\% for F1-score as the best performance by random forest.

Grondahl and Asokan \cite{Grondahl2019} surveyed deception detection from textual format and provided their arguments for different questions in this subject. 
Based on their findings, linguistic features are not enough for textual deception detection and it is better to also consider approaches, which are based on content comparison. They also discussed adversarial ``stylometry,'' the identification of authors based on writing style.

\section{Ensemble Learning and Classifiers}
\label{sec:compared}

This section presents background information regarding ensemble learning and the classifiers examined in this paper.  

\subsection{Ensemble Learning} 

Using ensemble learning, the performance of poorly performing classifiers can be improved by creating, training, and combining the output of multiple classifiers and thus result in a more robust classification. There are three main approaches for developing an ensemble learner \cite{Zhou2009}: 

\begin{itemize}
    \item {\it Boosting}, often uses homogeneous-base models trained sequentially;
    \item {\it Bagging}, which often uses homogeneous-base models trained in parallel; and
    \item {\it Stacking}, which uses mostly heterogeneous-base models trained in parallel and combined using a meta-model.
\end{itemize}

By averaging (or voting) the outputs produced by the pool of classifiers, especially for small dataset, ensemble methods provide better predictions and avoid overfitting. Another reason that contributes to the better performance of ensemble learning is its ability in escaping from the local minimum. By using multiple models, the search space becomes wider and the chance for finding a better output becomes higher~\cite{Sagi2018}. Ensemble learning may solve some machine learning limitations and challenges such as class imbalance, concept drift (change of features or labels over time), and curse of dimensionality.



\subsection{Machine/Deep Learning Classifiers}


We conducted our experiments using four classifiers: Decision Tree, Random Forest, Support Vector Machines (SVMs), Extreme Gradient-Boosting Trees (XGBT), and Multilayer Perceptron.

\section{Experimental Setup}
\label{sec:setup}

\subsection{Data Collection: ``Restaurant Dataset''}

We developed our own fake reviews dataset (the ``{\it Restaurant Dataset},'') in order to support research community with new dataset and thus help in addressing threats to the external validity of the experiments. The dataset consists of reviews of three restaurants, each of which has equal numbers of fake and real reviews, as well as positive and negative ones. We targeted restaurants that offered the same type of food, in order to avoid external bias. To collect these reviews, four undergraduate students wrote fake reviews (positive and negative) for three restaurants, each one paragraph in length. In addition, they collected real reviews for those three restaurants that were available from Google. To make sure that the real reviews were credible, we selected reviews from verified users. Finally, all the reviews and their labels were added to the dataset. The dataset includes 86 reviews of which 43 were fake reviews and 43 were real reviews, across all three restaurants. 

\subsection{Document Embedding for the ``Restaurant Dataset''}

For all classifiers in this work, we generated the features for our reviews using Doc2Vec \cite{le2014distributed}. Doc2Vec is an extension of Word2Vec. Similar to the way Word2Vec aims to represent words as vectors, Doc2Vec embeds documents as vectors \cite{le2014distributed}. In order to generate the document embedding, we preprocessed the dataset removing special characters, stop words, and punctuation marks.

\subsection{Classification Metrics}

We employed widely adopted classification metrics, such as accuracy, precision, recall, and $F_1$ score. An accuracy of $1.0$ indicates that the predicted labels and observed labels are identical. Precision is the ratio of true positives against all predicted positive labels. Recall is the ratio of predicted true positives against all observed true positive labels. The $F_1$ score is the harmonic mean of precision and recall, which yields a more informative performance metric when precision and recall have dissimilar values. Although we report accuracy and $F_1$, we determine overall model performance using accuracy, because our balanced dataset prevents accuracy to report misleading values.



\subsection{The Hyperparameter Optimization Algorithm}

We developed Python scripts for the chosen classifiers and used the Scikit-learn~
library and \textit{XGBoost} for the Extreme Gradient Boosting Trees~\cite{chen2015xgboost}. We carried out several experiments to identify the best hyperparameters for each classifier. 

Let $X$ and $y$ be the samples and their labels, respectively, $C$ be a classifier with $l$ hyperparameters to be tuned, and $H$ be a list where the $i^{th}$ element is either a probability distribution for the values of the $i^{th}$ hyperparameter of $C$, or a list of equiprobable categorical values for the hyperparameter. A randomized search takes hyperparameter values from $H$ for a fixed number of iterations, builds an instance of $C$ using them, and performs the classification. In our work, we executed the randomized search 100 times, saving the model estimated in each round for future comparison. Each model was also fitted using \textit{K}-fold cross-validation with $K = 10$ in each iteration. Once all iterations were completed, we selected the best model according to its accuracy, and returned both the model and its set of best hyperparameters. Our approach is defined in Algorithm \ref{algo:rand_search}. 
Algorithm \ref{algo:rand_search} also returns $X_{test}$ and $y_{test}$. These sets comprise unseen data and are used for the optimal classification and computing the final test errors.

\begin{algorithm}[t!]
    \caption{Hyperparameters selection via randomized search}
    \label{algo:rand_search}
    \begin{algorithmic}
        \Input
        \Desc{$X$}{Samples}; 
        \Desc{$y$}{Samples labels}
        \Desc{$H$}{Probability distributions for the $l$ hyperparameters}
        \Desc{$C$}{Classification model}
        \EndInput
        \Output
        \Desc{$M'$}{Best model}
        \Desc{$H'$}{Best set of hyperparameters}
        \Desc{$X_{test}$}{Test samples}
        \Desc{$y_{test}$}{Test samples labels}
        \EndOutput
        \vspace{5px}
        \vspace{5px}
    \end{algorithmic}
    \begin{algorithmic}[1]
        \State $X_{train}, y_{train}, X_{test}, y_{test} \gets $ split\_dataset$\left(X, y\right)$
        \State $M = \{\}$
        \For{$i \gets 1$ to $100$}
            \State $h_i \gets sample(H)$
            \State $Model = Fit-K-Fold-CV(X_{train}, y_{train}, C, h_i)$
            \State $M \gets M \cup Model$
        \EndFor
        \State $M' = GetBestModel(M)$
        \State $H' = GetParams(M')$
        \State \Return $M', H', X_{test}, y_{test}$
    \end{algorithmic}
      \vspace{-0.05in}
\end{algorithm}

The number of models estimated during the randomized search is limited to the number of iterations; however, fewer models are estimated in the event that the possible combinations of hyperparameters are fewer than the number of iterations. This procedure also prevents us from generating duplicated sets of hyperparameters.

The probability distribution over the values, or the possible categorical values, of the classifiers' hyperparameters, follows:

\begin{enumerate}
    \item Decision Tree
    \begin{itemize}
        \item {\it Max depth}: discrete uniform distribution over $[3, 5)$. This range was obtained empirically considering the size of our dataset, in order to prevent over-fitting.
        \item {\it Splitter}: best, random.
        \item {\it Criterion}: gini, entropy.
    \end{itemize}{}
    \item Random Forest
    \begin{itemize}
        \item {\it Number of weak classifiers}: discrete uniform distribution over $[100, 1000)$.
        \item {\it Max depth}: discrete uniform distribution over $[3, 5)$.
        \item {\it Criterion}: gini, entropy.
    \end{itemize}
    \item Support Vector Machine
    \begin{itemize}
        \item {\it Kernel}: RBF, linear, polynomial, sigmoid.
        \item C: continuous uniform over $[0.1, 1000)$.
        \item {\it Degree}: discrete uniform distribution over $[3, 10)$. Only considered when the kernel is polynomial.
        \item {\it Gamma ($\gamma$)}: scale, auto. Only considered when the kernel is RBF.
    \end{itemize}
    \item Extreme Gradient Boosting Trees
    \begin{itemize}
        \item {\it Number of stages}: discrete uniform distribution over $[100, 1000)$.
        \item {\it Max depth}: discrete uniform distribution over $[3, 5)$.
        \item {\it Gamma ($\gamma$)}: continuous uniform distribution over $[0, 10)$.
        \item {\it Learning rate ($\alpha$)}: continuous uniform distribution over $[0.01, 1)$.
    \end{itemize}
    \item Multilayer Perceptron
    \begin{itemize}
        \item {\it Maximum iterations}: uniformly sampled from the set $\{600, 800, 1000, 1200\}$.
        \item {\it Hidden layers size}: sampled from a discrete uniform distribution over $[2, 5)$. The number of units in each layer is uniformly sampled, once per layer, from $\{5, 10, 15, 20, 25, 30, 35, 40, 45, 50\}$.
        \item {\it Activation function}: ReLU, hyperbolic tangent, logistic sigmoid, identity function.
    \end{itemize}
\end{enumerate}

\section{Results and discussion}
\label{sec:results}

Table \ref{tab:all_hyper} shows the top set of hyperparameters that result in higher accuracy during the random search for Decision Tree, Random Forest, SVM, XGBT, and MLP, respectively. We report the five best sets of hyperparameters for each classifier, except for the Random Forest, which  has only three possible combinations of hyperparameters reporting a different accuracy. The accuracy and $F_1$ scores reported in Table \ref{tab:all_hyper} represent the mean of the estimates calculated during the 10-fold cross-validation for each model, using $X_{train}$ and $y_{train}$ of Algorithm \ref{algo:rand_search} to fit them, so they correspond to training errors.

\begin{table*}[h!]
    \caption{Best hyperparameters tuning (Training stage).}
    \label{tab:all_hyper}
    \centering
    \begin{tabular}{|c|c|c|c|c|c|c|c|c|}
    \hline
    \multicolumn{1}{|c|}{Classifier} &
    \multicolumn{4}{|c|}{Hyperparameters} &
      \multicolumn{1}{c|}{\multirow{2}{*}{\begin{tabular}[c]{@{}c@{}}Mean\\ Acc.\end{tabular}}} &
      \multicolumn{1}{c|}{\multirow{2}{*}{\begin{tabular}[c]{@{}c@{}}Std\\ Acc.\end{tabular}}} &
      \multicolumn{1}{c|}{\multirow{2}{*}{\begin{tabular}[c]{@{}c@{}}Mean\\ $F_1$\end{tabular}}} &
      \multicolumn{1}{c|}{\multirow{2}{*}{\begin{tabular}[c]{@{}c@{}}Std\\ $F_1$\end{tabular}}} \\ \cline{1-5}
      \multicolumn{1}{|c|}{Decision Tree} &
    \multicolumn{1}{|c|}{Criterion} &
      \multicolumn{1}{c|}{Max Depth} &
      \multicolumn{1}{c|}{Splitter} &
      \multicolumn{1}{c|}{--} &
      \multicolumn{1}{c|}{} &
      \multicolumn{1}{c|}{} &
      \multicolumn{1}{c|}{} &
      \multicolumn{1}{c|}{} \\ 
      \hline
&
    \multicolumn{1}{|c|}{gini} &
      \multicolumn{1}{c|}{3} &
      \multicolumn{1}{c|}{best} &
    \multicolumn{1}{c|}{--} &
      \multicolumn{1}{c|}{\textbf{0.796}} &
      \multicolumn{1}{c|}{0.097} &
      \multicolumn{1}{c|}{\textbf{0.769}} &
      \multicolumn{1}{c|}{0.123} \\ 
      \hline
      &
    \multicolumn{1}{|c|}{gini} &
      \multicolumn{1}{c|}{4} &
      \multicolumn{1}{c|}{best} &
          \multicolumn{1}{c|}{--} &
      \multicolumn{1}{c|}{0.772} &
      \multicolumn{1}{c|}{0.115} &
      \multicolumn{1}{c|}{0.751} &
      \multicolumn{1}{c|}{0.137} \\ 
      \hline
      &
    \multicolumn{1}{|c|}{gini} &
      \multicolumn{1}{c|}{3} &
      \multicolumn{1}{c|}{random} &
            \multicolumn{1}{c|}{--} &
      \multicolumn{1}{c|}{0.694} &
      \multicolumn{1}{c|}{0.096} &
      \multicolumn{1}{c|}{0.620} &
      \multicolumn{1}{c|}{0.136} \\ 
      \hline
      &
    \multicolumn{1}{|c|}{entropy} &
      \multicolumn{1}{c|}{3} &
      \multicolumn{1}{c|}{random} &
            \multicolumn{1}{c|}{--} &
      \multicolumn{1}{c|}{0.683} &
      \multicolumn{1}{c|}{0.092} &
      \multicolumn{1}{c|}{0.598} &
      \multicolumn{1}{c|}{0.123} \\ 
      \hline
      &
    \multicolumn{1}{|c|}{entropy} &
      \multicolumn{1}{c|}{4} &
      \multicolumn{1}{c|}{random} &
            \multicolumn{1}{c|}{--} &
      \multicolumn{1}{c|}{0.657} &
      \multicolumn{1}{c|}{0.095} &
      \multicolumn{1}{c|}{0.576} &
      \multicolumn{1}{c|}{0.148} \\ 
      \hline
      \hline
   \multicolumn{1}{|c|}{} &
    \multicolumn{4}{|c|}{Hyperparameters} &
      \multicolumn{1}{c|}{\multirow{2}{*}{\begin{tabular}[c]{@{}c@{}}Mean\\ Acc.\end{tabular}}} &
      \multicolumn{1}{c|}{\multirow{2}{*}{\begin{tabular}[c]{@{}c@{}}Std\\ Acc.\end{tabular}}} &
      \multicolumn{1}{c|}{\multirow{2}{*}{\begin{tabular}[c]{@{}c@{}}Mean\\ $F_1$\end{tabular}}} &
      \multicolumn{1}{c|}{\multirow{2}{*}{\begin{tabular}[c]{@{}c@{}}Std\\ $F_1$\end{tabular}}} \\ \cline{1-5}
          \multicolumn{1}{|c|}{Random Forest} &
    \multicolumn{1}{|c|}{Criterion} &
      \multicolumn{1}{c|}{Max Depth} &
      \multicolumn{1}{c|}{N Estimators} &
            \multicolumn{1}{c|}{--} &
      \multicolumn{1}{c|}{} &
      \multicolumn{1}{c|}{} &
      \multicolumn{1}{c|}{} &
      \multicolumn{1}{c|}{} \\ 
      \hline
&
    \multicolumn{1}{|c|}{entropy} &
      \multicolumn{1}{c|}{3} &
      \multicolumn{1}{c|}{931} &
            \multicolumn{1}{c|}{--} &
      \multicolumn{1}{c|}{\textbf{0.715}} &
      \multicolumn{1}{c|}{0.115} &
      \multicolumn{1}{c|}{\textbf{0.673}} &
      \multicolumn{1}{c|}{0.158} \\ 
      \hline
      &
    \multicolumn{1}{|c|}{gini} &
      \multicolumn{1}{c|}{4} &
      \multicolumn{1}{c|}{827} &
            \multicolumn{1}{c|}{--} &
      \multicolumn{1}{c|}{0.703} &
      \multicolumn{1}{c|}{0.118} &
      \multicolumn{1}{c|}{0.664} &
      \multicolumn{1}{c|}{0.156} \\ 
      \hline
    &
    \multicolumn{1}{|c|}{gini} &
      \multicolumn{1}{c|}{3} &
      \multicolumn{1}{c|}{659} &
            \multicolumn{1}{c|}{--} &
      \multicolumn{1}{c|}{0.690} &
      \multicolumn{1}{c|}{0.131} &
      \multicolumn{1}{c|}{0.658} &
      \multicolumn{1}{c|}{0.168} \\ 
      \hline
          \hline
    \multicolumn{1}{|c|}{} &
    \multicolumn{4}{|c|}{Hyperparameters} &
      \multicolumn{1}{c|}{\multirow{2}{*}{\begin{tabular}[c]{@{}c@{}}Mean\\ Acc.\end{tabular}}} &
      \multicolumn{1}{c|}{\multirow{2}{*}{\begin{tabular}[c]{@{}c@{}}Std\\ Acc.\end{tabular}}} &
      \multicolumn{1}{c|}{\multirow{2}{*}{\begin{tabular}[c]{@{}c@{}}Mean\\ $F_1$\end{tabular}}} &
      \multicolumn{1}{c|}{\multirow{2}{*}{\begin{tabular}[c]{@{}c@{}}Std\\ $F_1$\end{tabular}}} \\ \cline{1-5}
    \multicolumn{1}{|c|}{Support Vector Machine} &
    \multicolumn{1}{|c|}{C} &
      \multicolumn{1}{c|}{Degree} &
      \multicolumn{1}{c|}{$\gamma$} &
      \multicolumn{1}{c|}{Kernel} &
      \multicolumn{1}{c|}{} &
      \multicolumn{1}{c|}{} &
      \multicolumn{1}{c|}{} &
      \multicolumn{1}{c|}{} \\ 
      \hline
      &
    \multicolumn{1}{|c|}{21.832} &
      \multicolumn{1}{c|}{6} &
      \multicolumn{1}{c|}{scale} &
      \multicolumn{1}{c|}{RBF} &
      \multicolumn{1}{c|}{\textbf{0.726}} &
      \multicolumn{1}{c|}{0.116} &
      \multicolumn{1}{c|}{\textbf{0.699}} &
      \multicolumn{1}{c|}{0.130} \\ 
      \hline
      &
    \multicolumn{1}{|c|}{38.581} &
      \multicolumn{1}{c|}{7} &
      \multicolumn{1}{c|}{scale} &
      \multicolumn{1}{c|}{RBF} &
      \multicolumn{1}{c|}{0.704} &
      \multicolumn{1}{c|}{0.075} &
      \multicolumn{1}{c|}{0.669} &
      \multicolumn{1}{c|}{0.092} \\ 
      \hline
      &
    \multicolumn{1}{|c|}{33.202} &
      \multicolumn{1}{c|}{5} &
      \multicolumn{1}{c|}{scale} &
      \multicolumn{1}{c|}{RBF} &
      \multicolumn{1}{c|}{0.693} &
      \multicolumn{1}{c|}{0.087} &
      \multicolumn{1}{c|}{0.662} &
      \multicolumn{1}{c|}{0.102} \\ 
      \hline
      &
    \multicolumn{1}{|c|}{5.404} &
      \multicolumn{1}{c|}{4} &
      \multicolumn{1}{c|}{scale} &
      \multicolumn{1}{c|}{RBF} &
      \multicolumn{1}{c|}{0.682} &
      \multicolumn{1}{c|}{0.083} &
      \multicolumn{1}{c|}{0.613} &
      \multicolumn{1}{c|}{0.117} \\ 
      \hline
      &
    \multicolumn{1}{|c|}{4.843} &
      \multicolumn{1}{c|}{9} &
      \multicolumn{1}{c|}{scale} &
      \multicolumn{1}{c|}{RBF} &
      \multicolumn{1}{c|}{0.671} &
      \multicolumn{1}{c|}{0.091} &
      \multicolumn{1}{c|}{0.589} &
      \multicolumn{1}{c|}{0.144} \\ 
      \hline
          \hline
              \multicolumn{1}{|c|}{} &
    \multicolumn{4}{|c|}{Hyperparameters} &
      \multicolumn{1}{c|}{\multirow{2}{*}{\begin{tabular}[c]{@{}c@{}}Mean\\ Acc.\end{tabular}}} &
      \multicolumn{1}{c|}{\multirow{2}{*}{\begin{tabular}[c]{@{}c@{}}Std\\ Acc.\end{tabular}}} &
      \multicolumn{1}{c|}{\multirow{2}{*}{\begin{tabular}[c]{@{}c@{}}Mean\\ $F_1$\end{tabular}}} &
      \multicolumn{1}{c|}{\multirow{2}{*}{\begin{tabular}[c]{@{}c@{}}Std\\ $F_1$\end{tabular}}} \\ \cline{1-5}
          \multicolumn{1}{|c|}{E.G.Boosting Trees} &
    \multicolumn{1}{|c|}{$\gamma$} &
      \multicolumn{1}{c|}{$\alpha$} &
      \multicolumn{1}{c|}{M. depth} &
      \multicolumn{1}{c|}{N Estimators} &
      \multicolumn{1}{c|}{} &
      \multicolumn{1}{c|}{} &
      \multicolumn{1}{c|}{} &
      \multicolumn{1}{c|}{} \\ 
      \hline
      &
    \multicolumn{1}{|c|}{9.08} &
      \multicolumn{1}{c|}{0.035} &
      \multicolumn{1}{c|}{4} &
      \multicolumn{1}{c|}{421} &
      \multicolumn{1}{c|}{\textbf{0.783}} &
      \multicolumn{1}{c|}{0.094} &
      \multicolumn{1}{c|}{\textbf{0.726}} &
      \multicolumn{1}{c|}{0.151} \\ 
      \hline
      &
    \multicolumn{1}{|c|}{8.214} &
      \multicolumn{1}{c|}{0.036} &
      \multicolumn{1}{c|}{3} &
      \multicolumn{1}{c|}{895} &
      \multicolumn{1}{c|}{0.772} &
      \multicolumn{1}{c|}{0.071} &
      \multicolumn{1}{c|}{0.714} &
      \multicolumn{1}{c|}{0.134} \\ 
      \hline
      &
    \multicolumn{1}{|c|}{8.977} &
      \multicolumn{1}{c|}{0.409} &
      \multicolumn{1}{c|}{4} &
      \multicolumn{1}{c|}{791} &
      \multicolumn{1}{c|}{0.761} &
      \multicolumn{1}{c|}{0.078} &
      \multicolumn{1}{c|}{0.706} &
      \multicolumn{1}{c|}{0.134} \\ 
      \hline
      &
    \multicolumn{1}{|c|}{7.144} &
      \multicolumn{1}{c|}{0.231} &
      \multicolumn{1}{c|}{3} &
      \multicolumn{1}{c|}{544} &
      \multicolumn{1}{c|}{0.750} &
      \multicolumn{1}{c|}{0.082} &
      \multicolumn{1}{c|}{0.697} &
      \multicolumn{1}{c|}{0.140} \\ 
      \hline
      &
    \multicolumn{1}{|c|}{2.913} &
      \multicolumn{1}{c|}{0.183} &
      \multicolumn{1}{c|}{3} &
      \multicolumn{1}{c|}{406} &
      \multicolumn{1}{c|}{0.740} &
      \multicolumn{1}{c|}{0.097} &
      \multicolumn{1}{c|}{0.702} &
      \multicolumn{1}{c|}{0.156} \\ 
      \hline
      \hline
      
     \multicolumn{1}{|c|}{} &
    \multicolumn{4}{|c|}{Hyperparameters} &
      \multicolumn{1}{c|}{\multirow{2}{*}{\begin{tabular}[c]{@{}c@{}}Mean\\ Acc.\end{tabular}}} &
      \multicolumn{1}{c|}{\multirow{2}{*}{\begin{tabular}[c]{@{}c@{}}Std\\ Acc.\end{tabular}}} &
      \multicolumn{1}{c|}{\multirow{2}{*}{\begin{tabular}[c]{@{}c@{}}Mean\\ $F_1$\end{tabular}}} &
      \multicolumn{1}{c|}{\multirow{2}{*}{\begin{tabular}[c]{@{}c@{}}Std\\ $F_1$\end{tabular}}} \\ \cline{1-5}
          \multicolumn{1}{|c|}{Multilayer Perceptron} &
    \multicolumn{1}{|c|}{Max iterations} &
      \multicolumn{1}{c|}{Hidden layers \& units} &
      \multicolumn{1}{c|}{Activation Function} &
            \multicolumn{1}{c|}{--} &
      \multicolumn{1}{c|}{} &
      \multicolumn{1}{c|}{} &
      \multicolumn{1}{c|}{} &
      \multicolumn{1}{c|}{} \\ 
      \hline
&
    \multicolumn{1}{|c|}{1000} &
      \multicolumn{1}{c|}{(35, 40, 20, 5)} &
      \multicolumn{1}{c|}{relu} &
            \multicolumn{1}{c|}{--} &
      \multicolumn{1}{c|}{\textbf{0.680}} &
      \multicolumn{1}{c|}{0.100} &
      \multicolumn{1}{c|}{\textbf{0.685}} &
      \multicolumn{1}{c|}{0.082} \\ 
      \hline
      &
    \multicolumn{1}{|c|}{400} &
      \multicolumn{1}{c|}{(5, 15, 45, 10)} &
      \multicolumn{1}{c|}{tanh} &
            \multicolumn{1}{c|}{--} &
      \multicolumn{1}{c|}{0.671} &
      \multicolumn{1}{c|}{0.120} &
      \multicolumn{1}{c|}{0.682} &
      \multicolumn{1}{c|}{0.107} \\ 
      \hline
    &
    \multicolumn{1}{|c|}{800} &
      \multicolumn{1}{c|}{(40, 10, 20, 30)} &
      \multicolumn{1}{c|}{relu} &
            \multicolumn{1}{c|}{--} &
      \multicolumn{1}{c|}{0.669} &
      \multicolumn{1}{c|}{0.146} &
      \multicolumn{1}{c|}{0.651} &
      \multicolumn{1}{c|}{0.168} \\ 
      \hline
     &
     \multicolumn{1}{|c|}{600} &
      \multicolumn{1}{c|}{(50, 5, 15, 45)} &
      \multicolumn{1}{c|}{relu} &
            \multicolumn{1}{c|}{--} &
      \multicolumn{1}{c|}{0.668} &
      \multicolumn{1}{c|}{0.124} &
      \multicolumn{1}{c|}{0.647} &
      \multicolumn{1}{c|}{0.154} \\ 
      \hline
      &
     \multicolumn{1}{|c|}{600} &
      \multicolumn{1}{c|}{(10, 30, 20, 35)} &
      \multicolumn{1}{c|}{tanh} &
            \multicolumn{1}{c|}{--} &
      \multicolumn{1}{c|}{0.667} &
      \multicolumn{1}{c|}{0.147} &
      \multicolumn{1}{c|}{0.678} &
      \multicolumn{1}{c|}{0.135} \\ 
      \hline
    \end{tabular}
      \vspace{-0.05in}
\end{table*}

As Table \ref{tab:all_hyper} shows, the Decision Tree has the highest training accuracy, with 79.6\%, where the ``best" splitter and the Gini criterion for information gain yield the highest metrics by a margin of 7.8\%. In this model, the value of maximum tree depth does not seem to make a difference in the performance.

XGBT follows in accuracy with 78.3\%. The metrics for XGBT suggest  smaller values for the learning rate ($\alpha$), in addition to higher values of the regularization coefficient ($\gamma$), generate better accuracy and $F_1$ scores. Moreover, the two top results show the trade-off between maximum tree depth and the number of estimators in the ensemble, where high performance can be achieved by increasing the maximum depth with fewer numbers of estimators, or maintaining a lower depth with more estimators. 

Next, SVMs report 72.6\% accuracy. The RBF kernel in each set of hyperparameters suggests that the Doc2Vec representation of our dataset benefits from the non-linear mapping of the RBF kernel; this finding agrees with the high values for $C$, which controls the width of the soft margin in the SVM restricting samples within the margin in non-linearly separable data. $\gamma$ is set to ``scale" in all results, which shows that considering the variance of all our samples, in addition with the number of features to calculate $\gamma$, leads to a better accuracy. The Degree is ignored by the model, as it is used only when the kernel function is polynomial.

Random Forest reports 71.5\% accuracy in training stage. The results for Random Forest show that a relatively high number of estimators achieves high accuracy independent of the maximum depth of the tree. This is consistent with the improvement in the  performance of the weak classifier (decision tree) according to the number of estimators \cite{breiman2001random}.

Finally, the MLP achieves 68\% accuracy in training stage. As expected, the most important factor of the performance of the network is its architecture. Each value of the ``Hidden layers \& units" is a tuple $t$ of positive integers, where $size(t) = \# \text{ of hidden layers}$, and the $i$-th element of $t$ corresponds the number of neurons in the $i$-th hidden layer. The five results were generated using an MLP with 4 hidden layers, the maximum number possible in our experiments. The number of iterations and number of neurons per layer do not seem to affect the performance of the MLP, as the values for these hyperparameters range over all possible values. Regarding the activation function, ReLu and hyperbolic tangent were chosen over the rest; this agrees with the choice of the RBF kernel for the SVMs, as the ReLu and hyperbolic tangent functions allow the model to generalize in a non-linear way.

\begin{table}[]
    \caption{Training and test errors for the classifiers.}
    \label{tab:train_test_error}
    \centering
    \begin{tabular}{ccc}
    \hline
    \multicolumn{1}{|c|}{\bf Classifier} &
      \multicolumn{1}{c|}{\bf Training Error} &
      \multicolumn{1}{c|}{\bf Test Error} \\ \hline
    \multicolumn{1}{|c|}{Decision Tree} & \multicolumn{1}{c|}{0.102} & \multicolumn{1}{c|}{0.455} \\ \hline
    \multicolumn{1}{|c|}{Random Forest} & \multicolumn{1}{c|}{0.057} & \multicolumn{1}{c|}{0.318}\\ \hline
    \multicolumn{1}{|c|}{SVM}           & \multicolumn{1}{c|}{0.239} & \multicolumn{1}{c|}{0.409}\\ \hline
    \multicolumn{1}{|c|}{XGBT}          & \multicolumn{1}{c|}{0.182} & \multicolumn{1}{c|}{0.364}\\ \hline
    \multicolumn{1}{|c|}{MLP}          & \multicolumn{1}{c|}{0.0} & \multicolumn{1}{c|}{0.318}\\ \hline
    \multicolumn{1}{|c|}{Bagging Ensemble (SVM)} &
      \multicolumn{1}{c|}{0.227} &
      \multicolumn{1}{c|}{0.318} \\ \hline
    \multicolumn{1}{|c|}{Adaboost Ensemble (SVM)} &
      \multicolumn{1}{c|}{0.250} &
      \multicolumn{1}{c|}{0.273} \\ \hline 
      \multicolumn{1}{|c|}{Bagging Ensemble (MLP)} &
      \multicolumn{1}{c|}{0.034} &
      \multicolumn{1}{c|}{0.318} \\ \hline 
      \multicolumn{1}{|c|}{Adaboost Ensemble (MLP)} &
      \multicolumn{1}{c|}{0.068} &
      \multicolumn{1}{c|}{\textbf{0.227}} \\ \hline 
    \end{tabular}
  \vspace{-0.1in}   
\end{table}

Table \ref{tab:train_test_error} shows the training and test errors for the classifiers. To report the training errors, we fit each classifier with the $X_{train}$ and $y_{train}$ dataset using the best set of hyperparameters presented in Table \ref{tab:train_test_error}. The test error was reported by making predictions with the classifiers using the $X_{test}$ and $y_{test}$ dataset. $X_{train}$, $y_{train}$, $X_{test}$, and $y_{test}$ correspond to the datasets described in Algorithm \ref{algo:rand_search}.

We trained an ensemble of SVMs and MLPs fitted with their respective best hyperparameters. We trained both ensembles using the bagging and adaboost techniques, with the number of estimators ranging from 2 to 22 in steps of 2. Table \ref{tab:train_test_error} includes the lowest test error for the four aforementioned ensembles. For both SVM and MLP, the adaboost ensembles produce the best test error, with 0.273 for SVMs (16, 18, and 20 estimators) and 0.227 for MLPs (4 estimators). 

\begin{figure}[t!]
  \centering
  \includegraphics[width=3.5in]{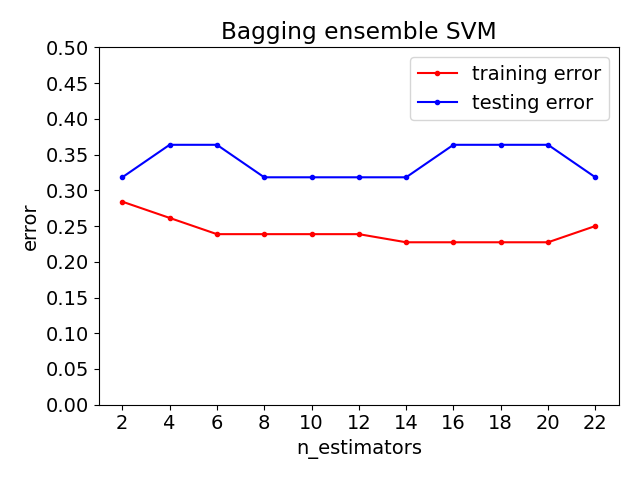}
  \caption{Training and test error for different number of estimators in the bagging ensemble for SVMs.}
  \label{fig:bag-svm}
  \vspace{-0.1in}
\end{figure}

\begin{figure}[t!]
  \centering
  \includegraphics[width=3.5in]{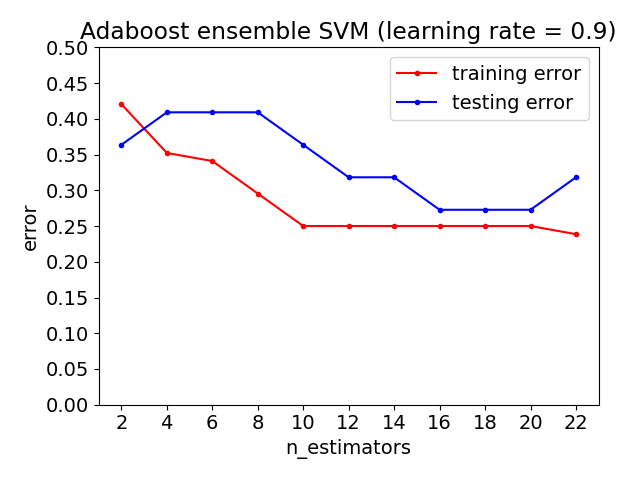}
  \caption{Training and test error for different number of estimators in the adaboost ensemble for SVMs.}
  \label{fig:ada-svm}
    \vspace{-0.19in}
\end{figure}

\begin{figure}[t!]
  \centering
  \includegraphics[width=3.5in]{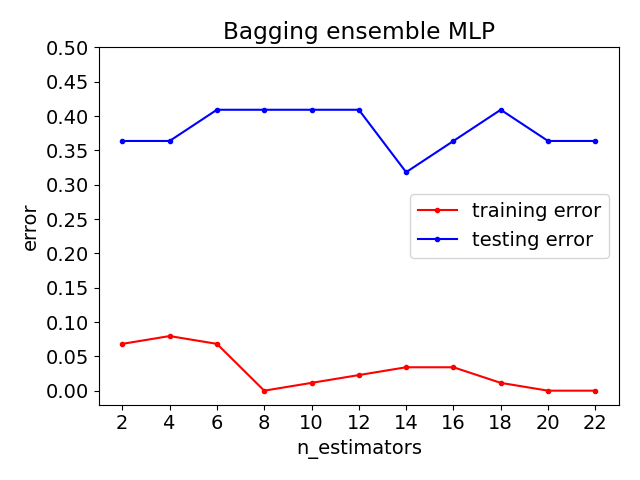}
  \caption{Training and test error for different number of estimators in the bagging ensemble for MLPs.}
  \label{fig:bag-mlp}
    \vspace{-0.1in}
\end{figure}

\begin{figure}[t!]
  \centering
  \includegraphics[width=3.5in]{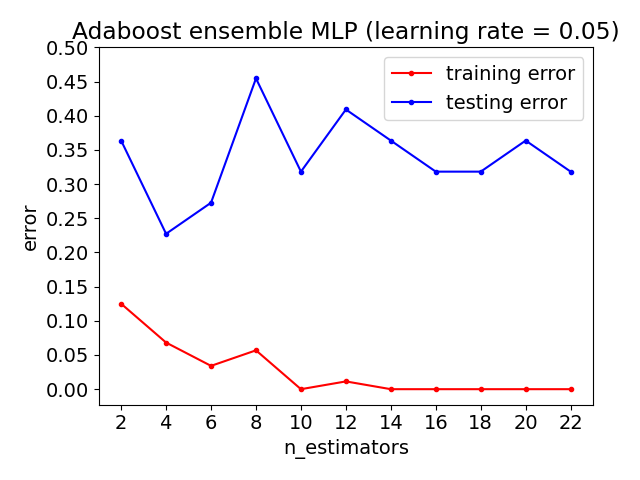}
  \caption{Training and test error for different number of estimators in the adaboost ensemble for MLPs.}
  \label{fig:ada-mlp}
    \vspace{-0.19in}
\end{figure}

Figures \ref{fig:bag-svm} and \ref{fig:ada-svm} show the training and testing errors for SVMs using bagging and adaboost, respectively. Figures \ref{fig:bag-mlp} and \ref{fig:ada-mlp} show the training and testing errors for MLPs using bagging and adaboost, respectively. The learning rates for adaboost were obtained empirically. The results of the ensemble of SVMs and MLPs suggest that the classification of Doc2Vec document embedding for the restaurant dataset benefits from adaboost over bagging. This finding might be because adaboost fits the estimators sequentially using the whole dataset, adjusting the ensemble to fit it in the best way; whereas bagging fits the estimators with different datasets that were sampled using a bootstrap from the original one \cite{Zhou2009}. This finding is consistent with the overall trend of the training error in Figure \ref{fig:ada-svm}, where it decreases steadily as the number of estimators grows, and Figure  \ref{fig:bag-mlp}, where this decreasing trend is also present, although with a few jumps. 
It is worth noting that the MLP adaboost ensemble fits the training data perfectly with 14 or more estimators; however, the testing error trend is less obvious for MLP ensembles. This result might be due to the limited number of samples in our dataset, as neural networks require several samples to be trained properly. 

\section{Conclusion and Future Work}
\label{sec:conclusion}

In this work, we present a novel dataset of fake reviews, along with the results of the binary classification using machine/deep learning techniques. Additionally, we applied ensemble learning-based approaches using Random Forest, bagging, and adaboost ensembles, with SVMs and MLPs as weak classifiers with optimized hyperparameters. Our results show that, using document embedding from Doc2Vec and after hyperparameter optimization, stand-alone classifiers can achieve up to 68.2\% accuracy in the case of MLP. Ensemble learning-based classifiers achieve up to 77.3\% accuracy with the adaboost ensemble of MLPs. In every case, the ensemble of classifiers outperforms their respective base classifier, either Random Forest and XGBT with Decision Tree, or the bagging/adaboost ensembles with their respective SVMs or MLPs. Regarding the ensemble approach, adaboost seems to produce the most consistent results. As future work, we can utilize heuristics in order to find optimum values for the hyperparameters of the models. Moreover, we need to explore our method with other different datasets, and also expand the Restaurant dataset in order to make any statistical analysis based on this dataset meaningful. We also intend to perform the classification with another set of features that allows more interpretability than Doc2Vec's document embedding, and provides more linguistic insights into deceptive texts. The focus of this work was on analyzing static texts. In more challenging settings,  when there are interactions between two parties (e.g., chatting and lie detection), more sophisticated analysis would be needed to take into account the time dimension (i.e., a time series problem) such as Long Short-Term Memory \cite{Siami-NaminiTN18, Siami-NaminiTN19} and clustering using deep learning \cite{SN-AS2020}, reinforcement learning \cite{ChatterjeeN19}, and evidence theory \cite{ChatterjeeN18}.

  \vspace{-0.01in}
\section*{Acknowledgment}
Thanks to Pritish Ayer, Sagar Lamichhane, Omer Qureshi, and Pranaya Sharma for contributions to the Restaurant dataset creation. This research work is supported by National Science Foundation under Grant No: 1723765.

\bibliographystyle{IEEEtran}
\bibliography{IEEEfull,sample-base}

\end{document}